\documentclass{article}

\PassOptionsToPackage{numbers,sort}{natbib}
\usepackage[final,eandd]{neurips_2026}

\usepackage[utf8]{inputenc}
\usepackage[T1]{fontenc}
\usepackage{hyperref}
\usepackage{url}
\usepackage{booktabs}
\usepackage{amsfonts}
\usepackage{nicefrac}
\usepackage{microtype}
\usepackage{xcolor}
\usepackage{CJKutf8}

\usepackage{times}
\usepackage{amssymb}
\usepackage{amsmath}
\usepackage{graphicx}
\usepackage{pifont}
\usepackage{multicol}
\usepackage{multirow}
\usepackage{enumitem}
\usepackage{caption}
\usepackage{subcaption}
\usepackage{placeins}
\usepackage[english]{babel}
\newcommand{\toolname}{LIBERO-PeRM}
\newcommand{\suiteSinglePref}{LIBERO-SinglePref}
\newcommand{\suitePrefScale}{LIBERO-PrefScale}
\newcommand{\suitePrefConflict}{LIBERO-PrefConflict}
\newcommand{\suitePrefCompose}{LIBERO-PrefCompose}
\newcommand{\suiteComprePref}{LIBERO-ComprePref}
\newcommand{\tasksuitenames}{\suiteSinglePref{}, \suitePrefScale{}, \suitePrefConflict{}, \suitePrefCompose{}, \suiteComprePref{}}
\newcommand{\tasknumber}{312}
\newcommand{\prefMotion}{\textit{Motion Preference}}
\newcommand{\prefSkill}{\textit{Skill Preference}}
\newcommand{\prefUserProfile}{\textit{User Profile}}

\newcommand{\cmark}{\textcolor{green!70!black}{\ding{51}}}
\newcommand{\xmark}{\textcolor{red}{\ding{55}}}

\newif\ifshowoldtext
\showoldtexttrue
\newcommand{\old}[1]{\ifshowoldtext\textcolor{red}{#1}\fi}

\title{\toolname{}: Benchmarking Personalized \\ Robotic Manipulation}

\author{%
  {\color{black}\bfseries Zhixu Li$^{1,*,\dagger}$, Keqian Tang$^{1,*}$, Litian Gong$^{1,*}$, Jingyu Yao$^{2}$, Wenqian Zhang$^{1}$,} \\
  {\color{black}\bfseries Tianze Xu$^{5}$, Zehao Wang$^{3}$, Yuping Wang$^{4}$, Junge Zhang$^{3}$, and Jiachen Li$^{3,1,\ddagger}$} \\
  {\color{black}\small $^{1}$University of California, Riverside \quad $^{2}$Yale University} \quad
  {\color{black}\small $^{3}$Georgia Institute of Technology} \\
  {\color{black}\small $^{4}$University of Michigan, Ann Arbor} \quad
  {\color{black}\small $^{5}$University of Toronto} \\
  {\color{black}\small $^{*}$Equal Contribution \quad $^{\dagger}$Project Lead \quad $^{\ddagger}$Corresponding Author: jiachen\_li@gatech.edu}
}

\begin{document}

\maketitle

\begin{abstract}
Personalization is essential for general-purpose robots to transition into household environments and achieve true practicality. While general robotic manipulation has been extensively studied, personalized robotic manipulation (PeRM) lacks a controlled benchmark for separating task completion from preference satisfaction. We introduce \toolname{}, a benchmark for preference learning in embodied agents. \toolname{} studies five questions: 1) how policies follow preferences in zero-shot settings; 2) how they learn single preferences from demonstrations; 3) how efficiently they acquire preferences as demonstrations increase; 4) how learned preferences generalize across layouts and tasks; and 5) how policies compose preferences and behave under conflicts. We propose a controlled formulation and taxonomy of observable preferences in simulation, covering motion, skill, and user-profile choices. This formulation is restricted to language-expressible preferences that can be instantiated in LIBERO without changing the task goal, evaluated automatically, and reused across tasks. Building on LIBERO, our procedural pipeline generates paired demonstrations with and without preferences. We create five task suites (\tasknumber{} tasks) with high-quality demonstrations, rich metadata, and preference-satisfaction metrics. Experiments show that current VLA policies struggle with zero-shot personalization, while SFT makes preferences learnable. Mixed training achieves better no-preference retention than sequential adaptation and improves performance with contextual prompts compared with preference-only training, but can weaken motion-preference learning; preference generalization, decomposition, and conflict handling remain difficult. The code, datasets, and demos can be found on the project website: {\urlstyle{same}\url{https://libero-perm.github.io/}}.

\end{abstract}

\section{Introduction}

Household manipulation has been a long-standing goal in robotics and embodied AI, with the promise of improving daily life by enabling robots to perform chores from natural language instructions (e.g., ``put an apple, bread, and milk on my plate''). Recent progress in generalist robotic manipulation has been driven in large part by Vision-Language-Action (VLA) models~\cite{pi0black2024,amin2025pi,openvlakim2024,grootbjorck2025}, which show strong generalization across diverse multimodal inputs~\cite{vlaszhao2025,pi05intelligence2025} and heterogeneous embodiments~\cite{openxo2024,univlabu2025}. However, real-world deployment demands more than completing the nominal task: \textit{embodied agents must also execute tasks in ways that align with the preferences, habits, and constraints of individual users}, a requirement we refer to as \textbf{personalization}.

\begin{figure}[!t]
    \centering
    \includegraphics[width=\textwidth]{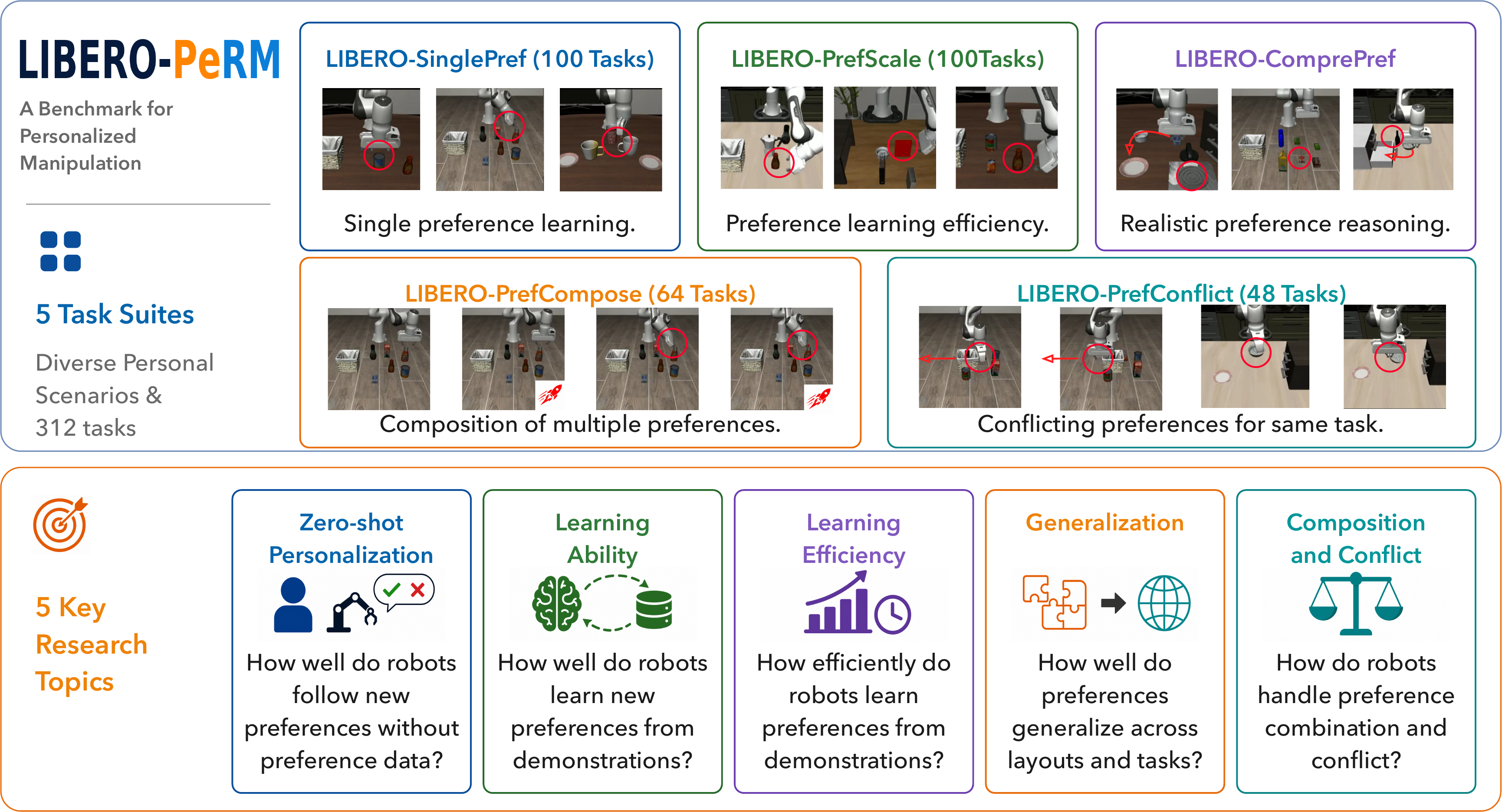}
    \captionsetup{width=\textwidth}
    \caption{Top: \toolname{} contains five procedurally generated task suites with pipeline-generated, human-reviewed demonstrations. Bottom: we investigate five key research questions in personalized manipulation.}
    \vspace{0.3cm}
    \label{fig:teaser}
\end{figure}

Preferences introduce diverse execution requirements for identical goals. For instance, placing a mug might require a specific handle orientation or a designated landing spot. Consequently, personalized manipulation must satisfy user-specific constraints on both \emph{what} strategies are selected and \emph{how} actions are performed. In this framework, preferences are treated as a learning dimension parallel to object-oriented task learning; the policy's objective is to maximize preference adherence within a single trajectory without compromising fundamental task success.

Nevertheless, progress on personalized robotic manipulation (PeRM) remains impeded by two major gaps.
First, PeRM lacks a shared formulation and standardized evaluation protocol across prior work.
Existing studies define personalization through diverse problem settings, including implicit preference alignment~\cite{feastjenamani2025,personalizingRLHFpoddar2024}, multi-user style adaptation~\cite{pecannemlekar2025}, user-specific object-receptacle pairing~\cite{llmpersonalizedhan2025}, and helpful action prediction~\cite{provox}; they also specify preferences through natural language~\cite{provox,alignbotzhaxizhuoma2024}, few user examples~\cite{feastjenamani2025}, explicit preference functions~\cite{graspvladeng2025}, human speech~\cite{vlaszhao2025}, or numerical preference vectors~\cite{pecannemlekar2025}.
Evaluation is similarly fragmented, spanning task success, user studies~\cite{provox,feastjenamani2025,pecannemlekar2025,alignbotzhaxizhuoma2024}, and preference-based scoring functions~\cite{pbarlwang2025,psharedliu2025}, making methods difficult to compare and failure modes hard to diagnose.
 Second, existing benchmarks typically emphasize high-level navigation or semantic selection over fine-grained manipulation preferences and multi-preference conflicts.
The closest preference-based RL benchmark~\cite{bpreflee2021} lacks robotic preference-learning datasets and multi-preference evaluation.
Other personalization benchmarks focus on object navigation or high-level receptacle selection~\cite{personalziliotto2025,tidybotwu2023}, while recent memory-based manipulation benchmarks~\cite{sam2actfang2025,starchen2025} do not distinguish task-instance memorization from genuine user-preference-driven personalization.

To address these gaps, we introduce a novel Personalized Robotic Manipulation Benchmark (\textbf{\toolname}), built using the popular LIBERO~\cite{liberoliu2023} framework. We elevate user preferences to a fundamental building block alongside tasks and layouts. Following the taxonomy of human preferences~\cite{surverypersonalizationzhang2024}, we define a controlled taxonomy of observable preferences in simulation, covering \prefMotion{} (trajectory constraints), \prefSkill{} (task strategies), and \prefUserProfile{} (semantic biases). The benchmark is restricted to preferences that can be expressed in language, instantiated without changing the original task goal, evaluated automatically, and reused across tasks. Within this scope, we provide 37 single-preference instances and 18 two-preference combinations.

\toolname{} features an extensible procedural generation pipeline that automatically instantiates tasks and their preference-conditioned variants within shared layouts. To support systematic training and evaluation, the pipeline simultaneously produces explicit preference satisfaction metrics and rich subtask-level metadata, including language instructions, goal images, and physical attributes.

For benchmarking, we construct five task suites with \tasknumber{} preference-conditioned tasks and pipeline-generated, human-reviewed demonstrations, each targeting a core capability in personalized manipulation (Figure~\ref{fig:teaser}, top).
\textit{\suiteSinglePref{}} evaluates single-preference learning; \textit{\suitePrefScale{}} evaluates scaling across data sizes, tasks, and layouts; \textit{\suitePrefConflict{}} evaluates preference-preference and task-preference conflicts; \textit{\suitePrefCompose{}} evaluates multi-preference composition; and \textit{\suiteComprePref{}} evaluates realistic preference-conditioned reasoning based on contextual language.
In \suitePrefScale{}, we further provide no-preference data, unseen layouts, and different tasks to support in-domain evaluation, task/layout generalization, and preference-level generalization over different tasks in unseen layouts.

We use \toolname{} to conduct an initial study of five key research topics in personalized manipulation (see bottom row of Figure~\ref{fig:teaser}): 1) how policies follow preferences in zero-shot settings; 2) how policies learn single preferences from demonstrations; 3) how efficiently policies acquire preferences as demonstrations increase; 4) how learned preferences generalize across layouts and tasks; and 5) how policies compose multiple preferences and behave under conflicts. These topics organize our experiments and highlight the core capabilities that robots must develop for reliable personalization.

\begin{table*}[t]
\centering

\caption{Comparison to existing manipulation benchmarks. Unlike prior benchmarks, \toolname{} targets personalized manipulation with large-scale preference-conditioned demonstrations and rich metadata, including subtask goals, subgoal images, trajectory statistics, and preference-satisfaction signals.}
\noindent %
\centering%
\resizebox{\textwidth}{!}{%

\renewcommand{\arraystretch}{1.2}
\setlength{\tabcolsep}{6pt}

\begin{tabular}{@{}l|c|c|c|c|c|l|l}
\hline
\textbf{Dataset} & \textbf{\# Traj.} & \textbf{\# Tasks} &
\textbf{Metadata} & \textbf{Pref. Data} &
\textbf{Pref. Metric} & \textbf{Targeted Ability} & \textbf{Data Collection} \\
\hline

ManiSkill2\cite{maniskill2gu2023} & 30k & 20  & \xmark & \xmark & \xmark &  Generalized, Sim-to-real Manipulation & TAMP \& MPC \& RL \\
RLBench\cite{rlbenchjames2020} & 2.7k & 100  & \xmark & \xmark & \xmark &  Multiple general  Manipulation types & Waypoints \& Planners \\
LIBERO\cite{liberoliu2023} & 6.5k & 130  & \xmark& \xmark  & \xmark & Lifelong Robot Manipulation & Teleoperation \\
CALVIN\cite{calvinmees2022} & 37.5k  & 34  & \cmark& \xmark  & \xmark &  Long-horizon Manipulation & Teleoperation \\
VLABench\cite{vlabenchzhang2025}& 1.6k  & 100  & \cmark& \xmark  & \xmark & Long-Horizon Reasoning Manipulation & Autonomous \\
RoboCasa\cite{robocasanasiriany2024} & 77k & 100  & \cmark & \xmark & \xmark &  Generalized, Sim-to-real Manipulation & Teleoperation \& Aug. \\
RoboTwin 2.0\cite{robotwinchen2025} & 100k & 50 &  \xmark& \xmark  & \xmark & Generalized, Sim-to-real Manipulation & Autonomous \\
InternVLA-M1\cite{internvla_m1chen2025} & 244k & 200 &  \cmark & \xmark & \xmark &   Long-horizon Manipulation  & Autonomous \\
InternData-A1\cite{interndata_a1tian2025} & 630k & 70 &  \cmark & \xmark & \xmark &  Multiple general Manipulation types & Autonomous \\
RoboMIND\cite{robomindwu2024} & 107k & 479 & \cmark & \xmark & \xmark & Multi-embodiment  Manipulation  & Teleoperation \\
RoboCerebra\cite{robocerebrahan2025}& 1k & 100 & \cmark & \xmark& \xmark  & Long-horizon Manipulation  & Teleoperation \\
LIBERO-OOD\cite{libero-ood}& - & 20 & \xmark & \xmark & \xmark & Generalized Manipulation & Derived from LIBERO \\
LIBERO-Plus\cite{libero_plusfei2025}& 22.4k & - & \cmark & \xmark & \xmark &  Generalized Manipulation  & Autonomous \\
LIBERO-Pro\cite{libero_prozhou2025}& -  & - & \xmark& \xmark  & \xmark & Generalized Manipulation  & - \\
LIBERO-X\cite{libero-x}& 2.5k & 600 & \cmark & \xmark & \xmark & Generalized Manipulation & Teleoperation \\
LIBERO-Para\cite{libero-para}& - & 10 & \xmark & \xmark & \xmark & Generalized Manipulation & Autonomous \\
\hline
\textbf{\toolname (Ours)} & 40.8k & \tasknumber{} &
 \cmark & \cmark &\cmark & \textbf{Personalized} Manipulation & Autonomous \\
\hline
\end{tabular}
}
\vspace{0.3cm}
\end{table*}
\label{benchmarkcompare}

Our contributions and findings are as follows:
\begin{itemize}[leftmargin=0.8em,labelsep=0.25em,itemsep=0.15em,topsep=0.25em]
    \item We introduce a controlled formulation and taxonomy of observable preferences in simulation, together with an automated pipeline that produces preference-conditioned demonstrations, rich metadata, and preference-satisfaction metrics.
    \item We construct five task suites, \tasksuitenames{}, comprising \tasknumber{} preference-conditioned tasks with pipeline-generated, human-reviewed demonstrations and evaluation protocols for controlled personalization studies.
    \item Extensive experiments show that current VLA policies struggle with zero-shot personalization, while SFT makes preferences learnable. Mixed training improves no-preference task success relative to sequential adaptation and contextual-prompt task success and preference satisfaction relative to preference-only LoRA, but weakens motion-preference learning; broader layout diversity and matched no-preference data also help, while preference generalization, decomposition, and execution under conflicts remain difficult.

\end{itemize}

\section{Related Work}
\textbf{Robotic Manipulation Benchmark.} 
Many benchmarks have been developed to evaluate different aspects of robotic manipulation, including cross-embodiment generalization~\cite{openxo2024,interndata_a1tian2025,robomindwu2024}, generalized manipulation~\cite{rlbenchjames2020,libero_prozhou2025,libero-para}, long-horizon reasoning~\cite{robocerebrahan2025,calvinmees2022,vlabenchzhang2025,internvla_m1chen2025}~\cite{zhang2025lamma,gupta2026scale}, and sim-to-real transfer~\cite{robocasanasiriany2024,robotwinchen2025,maniskill2gu2023}. Among them, LIBERO~\cite{liberoliu2023} has emerged as a widely adopted benchmark due to its high-quality demonstrations and standardized data generation pipeline, primarily focusing on lifelong learning and task generalization. However, existing benchmarks largely focus on task completion, without explicitly modeling user-specific preferences. In contrast, our \toolname~targets personalized manipulation with preference-aware task suites that disentangle task success from preference satisfaction. Table~\ref{benchmarkcompare} compares commonly used robotic manipulation benchmarks.

\textbf{Personalized Robotic Manipulation.}
Prior work uses personalization to cover heterogeneous settings, leaving the scope of personalized manipulation unclear.
Some studies frame it as trajectory-level preference alignment via human feedback, reward signals~\cite{personalizingRLHFpoddar2024,prismkirk2024}~\cite{ghosh2026roved,ghosh2025preference}, RL post-training~\cite{vlarllu2025,grapezhang2024,deeprltang2025}~\cite{11246039}, or implicit preference optimization~\cite{dporafailov2023direct,ktoethayarajh2024,droidkhazatsky2024}.
Others study multi-user style adaptation~\cite{pecannemlekar2025}~\cite{Wang_2026_CVPR}, object-receptacle pairing~\cite{llmpersonalizedhan2025}, or helpful action prediction~\cite{provox}.
In contrast, personalized manipulation requires an operational formulation for preference-conditioned physical action.
We address this gap with \toolname{}, which provides a controlled taxonomy of observable preferences in simulation and five preference-aware task suites targeting key personalization aspects.

\textbf{Vision-Language-Action (VLA) Models.}
Vision-Language-Action (VLA) models~\cite{pi05intelligence2025,openvlaoftkim2025fine,amin2025pi,grootbjorck2025,xvla,pi07}~\cite{gao2026neuro} have shown strong reasoning and generalization abilities by leveraging VLM backbones~\cite{palmedriess2023,rt2zitkovich2023,flamingoalayrac2022}, large-scale multimodal instruction datasets~\cite{visualliu2023,cocochen2015microsoft,cambriantong2024}, and diverse robotic datasets~\cite{openxo2024,droidkhazatsky2024,interndata_a1tian2025,agibotbu2025}~\cite{zhao2026axis}, alongside advances in VLM-based spatial reasoning~\cite{Fan_2026_CVPR}, policy guidance~\cite{11697343,gong2026autofocus}, and tactile- and force-aware robot learning~\cite{shan2026learning}.
However, existing benchmarks~\cite{liberoliu2023,calvinmees2022} mainly measure task success, leaving preference alignment underexplored despite its importance for home robots, where users may specify not only what task to perform but also how it should be executed.
In this work, we introduce \toolname{}, which evaluates five key personalization topics through preference-injected trajectories, natural-language preference instructions, and ground-truth preference satisfaction matrices.
For hierarchical VLAs~\cite{shi2025hi,yan2025rdd}, \toolname{} further provides subtask decompositions and subgoal-image annotations, covering the key signals needed to evaluate personalized manipulation.

\begin{figure}[!t]
    \centering
    \includegraphics[width=\textwidth,keepaspectratio]{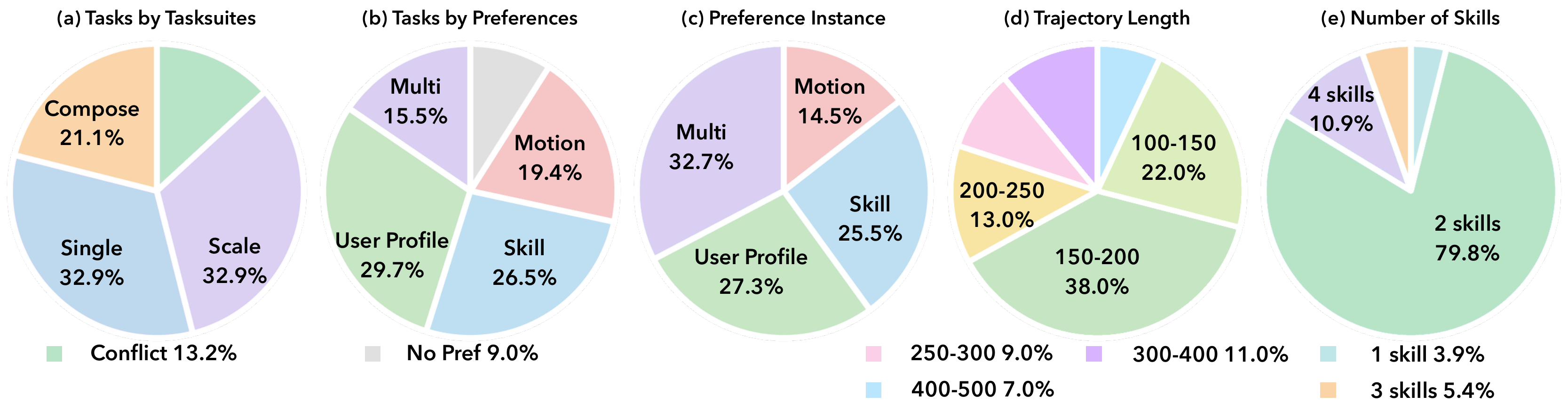}
    \caption{\toolname{} dataset statistics.
    (a) Distribution of tasks by task suite.
    (b) Distribution of tasks by preference type; No Preference denotes generated tasks from \toolname{}.
    (c) Distribution of preference instances by preference type.
    (d) Trajectory length distribution.
    (e) Distribution of skill numbers in a task.}
    \vspace{0.3cm}
    \label{fig:data_stat}
\end{figure}

\section{The \toolname{} Benchmark}
\label{sec:benchmark}

This section introduces the four components of \toolname{}. Section~\ref{pref_def} defines what preferences are; Section~\ref{sec:perm_pipeline} describes the procedural preference data generation pipeline; Section~\ref{sec:task_suites} defines the five task suites used for benchmarking; and Section~\ref{sec:eval_metrics} defines the evaluation methodologies.

\subsection{Preference Formulation}
\label{pref_def}
Preference describes the relative desirability among alternatives~\cite{scherer-emotion-measurement}. Personalized robotic manipulation, therefore, requires a policy to select, among feasible trajectories that complete a requested task, one that also satisfies the user's preference. A preference is a variable independent of the task goal: it constrains how the same goal is achieved rather than replacing the goal. Natural language is our interface for both variables, but this independence distinguishes the setting from merely redefining a language-conditioned task. We organize preferences by the part of execution they constrain: motion, skill, or subtask-level choices. Specifically, they are:
\begin{itemize}[leftmargin=1.2em,itemsep=0.25em,topsep=0.25em]
    \item \textbf{\prefMotion{}.} This type of preference constrains the physical properties of the trajectory while preserving task completion. For the task \textit{``put the mug on the table''}, the preference \textit{``avoid a particular area of the table''} induces a trajectory-level distance constraint, requiring the robot to keep the end-effector and mug away from that region during execution. We denote this kind of preference as $P_{\mathrm{motion}}$.

    \item \textbf{\prefSkill{}.} This type of preference alters the execution behavior of a specific skill $c$ while preserving task completion. For the task \textit{``put the mug on the table''}, the preference \textit{``orient the handle outward''} induces a pose-constrained placement strategy. When executing the placement skill, the robot must rotate the mug during placement so that the handle faces outward. We denote this kind of preference as $P_{\mathrm{skill}}$.

    \item \textbf{\prefUserProfile{}.} This type of preference uses a user profile or context as background knowledge to guide subtask-level semantic choices, such as selecting a target object or goal region. For the task \textit{``put the mug on the table''}, the user profile \textit{``prefers to protect table surfaces''} induces a target-selection constraint, causing the policy to choose a coaster as the support target while preserving the original task goal. We denote this kind of preference as $P_{\mathrm{profile}}$.
\end{itemize}

\begin{figure}[!t]
    \centering
    \includegraphics[width=\textwidth,keepaspectratio]{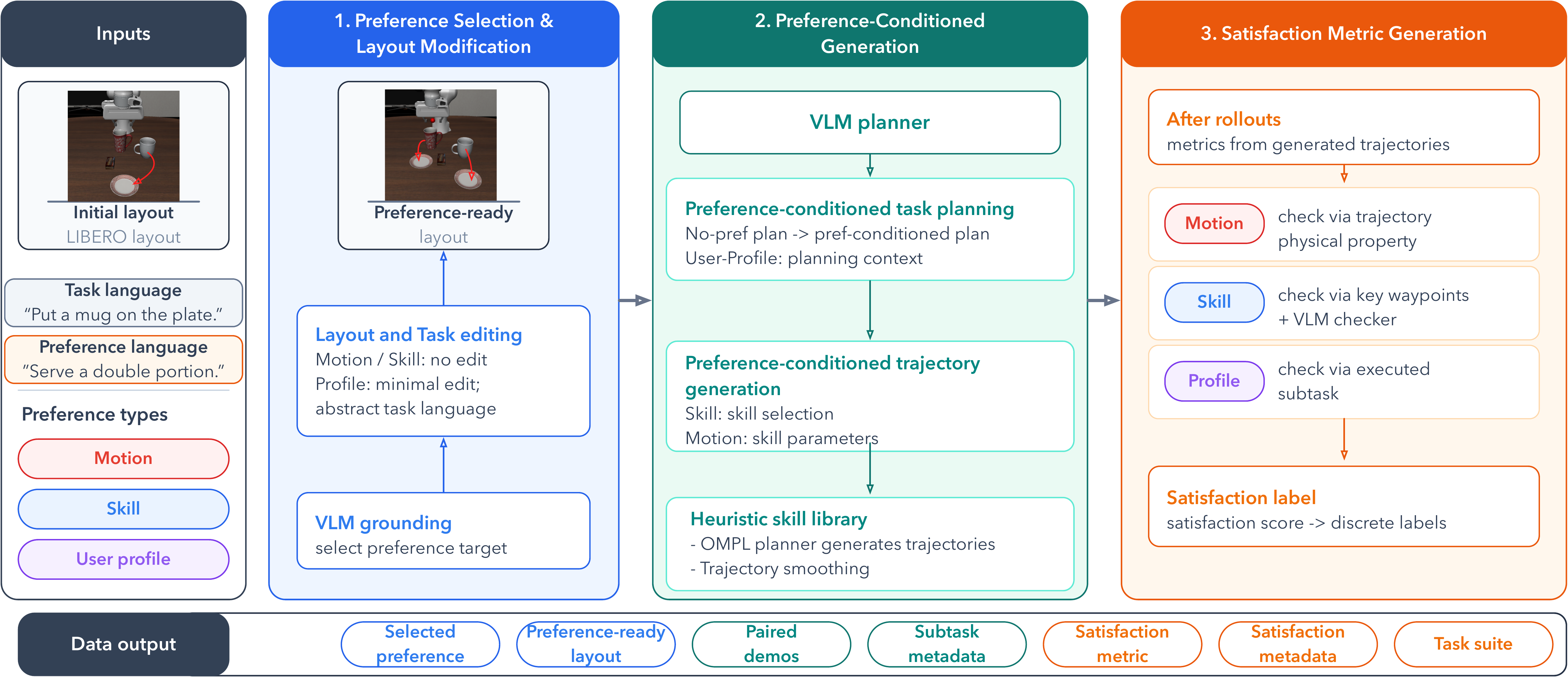}
    \caption{Data generation pipeline for personalized manipulation.
    Building on the LIBERO framework, the pipeline first generates preference-conditioned data, including preferences, conditions, subtasks, and trajectories, and then derives preference metrics and metadata.}
    \vspace{0.3cm}
    \label{fig:datagen}
\end{figure}

Together, these three types provide a controlled taxonomy of observable preferences in simulation, from user-profile conditioning and skill-specific execution to motion-level control. The current benchmark does not model implicit inference from interaction history, non-linguistic preference signals, or open-ended preferences that lack an automatic execution metric.

\textbf{Policy input and output.}
Let $\ell_{\text{task}}$ and $\ell_{\text{pref}}$ denote the task and preference language.
We concatenate them with text labels into the prompt \textit{``Task: \{task\_language\}. Preference: \{preference\_language\}.''}, denoted as $\ell_{\text{task}} \mathbin{\Vert} \ell_{\text{pref}}$.
At each timestep, the policy receives AgentView and Eye-in-Hand images, $I_t^{\mathrm{agent}}$ and $I_t^{\mathrm{hand}}$, and outputs an action:
$\pi_\theta(a_t \mid I_t^{\mathrm{agent}}, I_t^{\mathrm{hand}}, \ell_{\text{task}} \mathbin{\Vert} \ell_{\text{pref}})$.

\textbf{Preference Instances.}
With the three preference types defined above, we provide 37 single-preference instances, including 8 \prefMotion{} instances, 14 \prefSkill{} instances, and 15 \prefUserProfile{} instances. Appendix~\ref{app:preference_selection}, Table~\ref{tab:single_preference_catalog}, provides the complete list.
We further construct 18 multi-preference combinations from these instances.
Figure~\ref{fig:data_stat}(c) reports the resulting preference-instance distribution by type.
Figure~\ref{fig:data_stat}(b) shows the corresponding distribution in generated data tasks, including the no-preference tasks provided by \toolname{} for comparison. Detailed preference-selection criteria are provided in Appendix~\ref{app:preference_selection}.

\subsection{Preference-Conditioned Data Generation Pipeline}
\label{sec:perm_pipeline}
Building upon the LIBERO framework, \toolname{} provides an extensible procedural data-generation pipeline for preference learning and evaluation, as illustrated in Figure~\ref{fig:datagen}; model-assisted generation and validation details are given in Appendix~\ref{app:generation_details}.
The pipeline consists of three stages:
(1) Preference Selection and Layout Modification; (2) Preference-Conditioned Subtask and Trajectory Generation; and
(3) Satisfaction Metric Generation.

\textbf{Preference Selection and Layout Modification.}
\label{sec:IV_C}
The user first selects a preference instance and an initial LIBERO layout~\cite{liberoliu2023}.
Given these inputs, the pipeline uses a VLM~\cite{geminiteam2023} to minimally modify the layout so the selected preference can be operationalized while limiting changes to the original scene.
For \prefMotion{} and \prefSkill{} preferences, the layout and task language remain unchanged; the VLM only selects the target object or region, e.g., choosing the stove as the object to avoid for \textit{``avoid certain object''}.
For \prefUserProfile{} preferences, which may change the subtask plan, the pipeline makes minimal edits, such as adding a red object of the same category for a red-object preference, so preference-conditioned and no-preference trajectories can be generated in the same setting.
The task language is also abstracted to cover both settings, ensuring that behavioral differences are driven by the preference rather than unrelated scene changes.

\textbf{Preference-Conditioned Subtask and Trajectory Generation.}
After the layout and task language are fixed, the pipeline uses a VLM with the task language and preference language to generate both a no-preference subtask plan and a preference-conditioned subtask plan.
It then calls a library of pre-recorded heuristic skills to generate trajectory data.
To construct each heuristic skill, we first collect one human demonstration for the skill, extract key waypoints, and use an OMPL planner~\cite{ucan_The_Open_Motion_2012} with randomized initial states to generate many successful trajectories.
The generated trajectories are then smoothed; heuristic-skill parameters are manually offset-adjusted when needed. Empirically, we find more than 72\% of skills require no offset adjustment across any task, suggesting the VLM-based edit is already very effective.
Preferences are applied at different stages according to their type: \prefUserProfile{} affects subtask-plan generation via a prompt; \prefSkill{} affects skill selection, such as using \textit{pick\_handle} instead of a regular \textit{pick} skill; and \prefMotion{} affects skill-level trajectory generation, such as reducing OMPL-sampled waypoints for the \textit{``speed up''} preference.

\textbf{Satisfaction Metric Generation.}
After preference-conditioned trajectories are generated, the pipeline constructs explicit preference satisfaction metrics for each selected preference.
The form of each metric depends on the preference type.
For \prefMotion{}, we directly compute satisfaction functions from trajectory quantities, such as the distance between the robot trajectory and the target object for the preference \textit{``avoid certain object''}.
For \prefSkill{}, we evaluate key waypoints and use VLM-based image checks to verify skill-specific execution, such as whether the object is grasped by its handle.
For \prefUserProfile{}, we monitor the executed subtask plan, including which objects are manipulated, to verify that the plan adheres to the profile.

Because \prefUserProfile{} yields only discrete Satisfaction Scores, namely whether the profile-conditioned subtask is satisfied, whereas \prefMotion{} and \prefSkill{} can yield continuous Satisfaction Scores, we construct a mapping from Satisfaction Scores to Discrete Satisfaction Labels using trajectory statistics and human inspection; Appendix~\ref{app:human_validation} details the review protocol.
Specifically, each episode is mapped to a fixed four-level Discrete Satisfaction Label \(d \in \{0,1/4,3/4,1\}\), independent of the number of pipeline-generated subtasks.
A continuous Satisfaction Score, such as speed, is binned by trajectory statistics and then incorporated into this label.
Raw Satisfaction Scores provide denser feedback for training, while Discrete Satisfaction Labels provide a more interpretable evaluation signal at inference time.

\noindent\textbf{Initial Human Validation.}
To assess the real-world plausibility of the selected preferences and the human-likeness of the generated demonstrations, we conducted a survey with 15 participants using a 10-point scale.
Across all preference categories, the demonstrations received an average human-likeness score of 9.11, and the preferences received an average real-world plausibility score of 8.35.
These results provide initial human validation of the benchmark design; Appendix~\ref{app:human_validation} reports the protocol and category-level results.

\subsection{Task Suites}
\label{sec:task_suites}
Using the pipeline above, we generate five task suites: \suiteSinglePref{}, \suitePrefScale{}, \suitePrefConflict{}, \suitePrefCompose{}, and \suiteComprePref{}.
We follow LIBERO terminology and call the inseparable combination of a task instruction, scene layout, and task-goal condition a \emph{mission}. A \toolname{} \emph{task} pairs one mission with one preference condition, which may be a specified preference or no preference; the \tasknumber{} tasks include both preference-conditioned and no-preference tasks (see Figure~\ref{fig:data_stat}(b) for the proportions), with more than 100 trajectories per task on average and 40.8K rollout episodes in total. Randomized initial object positions, paraphrased instructions, and repeated executions increase rollout diversity but do not create additional tasks. The release prioritizes quality-controlled coverage over claiming the largest trajectory count; the procedural pipeline can extend it with more initial states and missions.
Figure~\ref{fig:data_stat}(a-c) shows distributions over preference instances, task suites, and preference types, including no-preference tasks for comparison.
Figure~\ref{fig:data_stat}(d) shows that episode horizons and trajectory fluency remain comparable to LIBERO.
Each suite isolates a core capability for personalized manipulation, as summarized in Figure~\ref{fig:teaser}.

\textbf{\suiteSinglePref{}.}
This suite evaluates single-preference learning and execution.
It contains 100 tasks and serves as the foundation for evaluating personalization. \textbf{\suitePrefScale{}.}
This suite evaluates how preference learning scales across demonstrations, tasks, and layouts.
It supports controlled analysis of preference acquisition and generalization across tasks and layouts.
It contains 100 tasks. \textbf{\suitePrefConflict{}.}
This suite evaluates policy behavior when preferences conflict with other preferences or with the task goal.
Preference-preference conflicts pair the same layout and task with mutually conflicting single preferences, while task-preference conflicts pair no-preference task data with task-incompatible preference data.
It contains 48 tasks. \textbf{\suitePrefCompose{}.}
This suite evaluates multi-preference learning and preference composition in a controlled two-preference setting.
For each aligned layout-task pair, it provides four conditions: no preference, \(P_A\), \(P_B\), and \(P_A + P_B\), enabling controlled comparisons for composition and multi-preference learning.
It contains 64 tasks. \textbf{\suiteComprePref{}.}
This suite rewrites prompts over tasks from the other suites so the preference must be inferred from contextual language rather than read from a separate explicit field.
It includes explicit task-preference mixtures, where key preference words are preserved, and implicit preference reasoning, where contextual cues imply the preference without naming it. For example, \textit{``Put both the alphabet soup and the cream cheese box in the basket. Preference: speed up''} becomes \textit{``Put both the alphabet soup and the cream cheese box in the basket. Guests will use them immediately.''} Similarly, \textit{``Pick the mug by its handle''} is conveyed as \textit{``The mug just came out of the microwave---the body is too hot to touch.''}
ComprePref reuses existing tasks, so its contextual prompts do not add to the 312-task count or the base-suite counts in Figure~\ref{fig:data_stat}(a). Appendix~\ref{app:comprepref_example} gives a worked example; Appendix~\ref{app:release} describes the release and maintenance plan.

\subsection{Evaluation}
\label{sec:eval_metrics}

\textbf{Task and Preference Metrics.}
We report two types of metrics: task success rate (SR), which measures task completion, and preference satisfaction (SA), which measures how well a preference is satisfied using the Discrete Satisfaction Label from Section~\ref{sec:perm_pipeline}.
The Discrete Satisfaction Label for episode \(n\) is \(d_n \in \{0,1/4,3/4,1\}\), and \(\mathbb{I}[G(\tau_n)]\) is its task-success indicator.
We compute \(\mathrm{SA}=\frac{1}{N}\sum_{n=1}^{N}\mathbb{I}[G(\tau_n)] d_n\), so preference satisfaction contributes only when the task succeeds.

\textbf{Conflict Policy.}
For a task-preference conflict, completing the original task takes priority. For preference-preference conflicts, SA is computed using the higher of the two preference satisfaction scores for each episode, with task-failure episodes contributing zero satisfaction. These scoring conventions probe execution under inconsistent instructions; they do not evaluate conflict detection, preference weighting, or when a policy should ask the user for guidance.

\textbf{Generalization.}
We evaluate in-domain, task-or-layout, and preference-level generalization using SR and SA.
In-domain evaluation keeps the training tasks, preferences, and layouts with small object-position perturbations.
Task-or-layout generalization changes either the task or layout, while preference-level generalization changes tasks under unseen layouts while keeping the preference unchanged.

\section{Experiments}
We conduct experiments on the five task suites introduced in Section~\ref{sec:task_suites} to evaluate zero-shot personalization, preference learning ability, preference learning efficiency, preference generalization, composition, and behavior under conflicts. Our experiments focus on the following research questions:

\noindent\textbf{Q1.} Can off-the-shelf policies follow preferences without preference-specific training?\\
\textbf{Q2.} Can supervised fine-tuning teach policies to satisfy preferences on seen tasks and layouts?\\
\textbf{Q3.} How does performance change as the number of demonstrations increases?\\
\textbf{Q4.} Do learned preferences generalize across different generalization levels?\\
\textbf{Q5.} Can a policy compose multiple preferences, and how does it behave under incompatible preferences?

\subsection{Experimental Setup}

We select six representative VLA policies for evaluation: OpenVLA-OFT~\cite{openvlaoftkim2025fine}, $\pi_{0}$-fast~\cite{pi0black2024}, $\pi_{0.5}$~\cite{pi05intelligence2025}, GR00T-N1.6~\cite{grootbjorck2025}, SmolVLA~\cite{shukor2025smolvla}, and UniVLA~\cite{univlabu2025}. For non-zero-shot experiments, we mainly use $\pi_{0.5}$, our strongest zero-shot baseline, as shown in Table~\ref{tab:main_results_zeroshot}. Training details are provided in Appendix~\ref{app:training_details}. For evaluation, we adopt SR and SA as defined in Section~\ref{sec:eval_metrics}. Table~\ref{tab:single_performance_sr} additionally reports the mean and standard deviation over 10 independent evaluation runs with fixed checkpoints; these error bars capture evaluation randomness rather than training-seed variation.

\subsection{Experimental Results}
\label{sec:experimental_results}

\noindent\textbf{Q1: Zero-shot Personalization.} Q1 asks whether off-the-shelf VLA policies can follow user preferences without preference-specific training.
Table~\ref{tab:main_results_zeroshot} evaluates six policies on the \suiteSinglePref{} task suite.
The no-preference column uses the original LIBERO task language with minimally edited \suiteSinglePref{} layouts.
\label{sec:rq1}
\begin{table*}[!t]
    \centering
    \footnotesize
    \renewcommand{\arraystretch}{1.15}
    \setlength{\tabcolsep}{4pt}
    \caption{Zero-shot preference on LIBERO-SinglePref. For each preference-conditioned column we report task success rate (SR, \%) and preference satisfaction (SA, \%); No-Pref reports SR only.}
    \label{tab:main_results_zeroshot}

    \makebox[\textwidth][c]{%
    \begin{tabular}{l|c|cc|cc|cc|cc}
        \hline
        \multirow{2}{*}{\textbf{Method}}
            & \textbf{No-Pref}
            & \multicolumn{2}{c|}{\textbf{All-Pref}}
            & \multicolumn{2}{c|}{\textbf{\prefMotion{}}}
            & \multicolumn{2}{c|}{\textbf{\prefSkill{}}}
            & \multicolumn{2}{c}{\textbf{\prefUserProfile{}}} \\ \cline{2-10}
            & SR & SR & SA & SR & SA & SR & SA & SR & SA \\ \hline
        OpenVLA-OFT~\cite{openvlaoftkim2025fine}       & 91.2 & 40.00 & 4.53 & 70.00 & 7.30 & 64.10 & 3.20 & 3.92 & 3.92 \\
        $\pi_{0}$-fast~\cite{pi0black2024}    & 86.88 & 51.25 & 30.37 & 80.00 & \textbf{27.60} & 73.08 & 18.27 & 17.65 & 17.65 \\
        \textbf{\boldmath$\pi_{0.5}$~\cite{pi05intelligence2025}}       & \textbf{97.6} & \textbf{66.67} & \textbf{37.92} & \textbf{97.50} & 24.38 & \textbf{100.00} & \textbf{25.00} & \textbf{19.12} & \textbf{19.12} \\
        GR00T-N1.6~\cite{grootbjorck2025} & 88.3 & 51.67 & 27.45 & 90.00 & 20.70 & 76.92 & 11.54 & 9.80 & 9.80 \\
        SmolVLA~\cite{shukor2025smolvla}     & 60.0 & 4.17 & 2.03 & 16.67 & 4.17 & 0.00 & 0.00 & 0.00 & 0.00 \\
        UniVLA~\cite{univlabu2025}            & 80.8 & 45.83 & 21.52 & 90.00 & 10.17 & 71.79 & 3.59 & 0.00 & 0.00 \\ \hline
    \end{tabular}
    }
\end{table*}

\noindent\textbf{\textit{Finding.}} Table~\ref{tab:main_results_zeroshot} shows that minimal layout edits from our pipeline preserve no preference zero-shot task performance, and existing SOTA VLA models fail to follow preferences in the preference-conditioned zero-shot setting.

Most policies maintain high no-preference SR, with $\pi_{0.5}$ reaching 97.6\%, indicating that the edited layouts preserve the LIBERO task distribution. Thus, low SA mainly reflects preference-following failure rather than layout-editing artifacts. Notably, $\pi_{0.5}$ reaches high SR on Motion and Skill Preference (97.50\% and 100.00\%), but low SA (24.38\% and 25.00\%), showing that it completes tasks while ignoring preference. Executing with a User Profile is harder since it changes the target object or semantic choice, reducing $\pi_{0.5}$ SR to 19.12\% and causing near-zero SR for weaker policies.

\noindent\textbf{Q2: Learning Ability.} Table~\ref{tab:single_performance_sr} reports in-domain and unseen-layout evaluation results under LoRA-based supervised fine-tuning (SFT).
The $\pi_{0.5}$-LoRA setting is used only on \suiteSinglePref{} to validate the effectiveness of our dataset.
\label{sec:rq2}

\begin{table*}[!t]
    \centering
    \footnotesize
    \renewcommand{\arraystretch}{1.15}
    \setlength{\tabcolsep}{2.3pt}
    \caption{Fine-tuned preference-learning performance on \toolname{}, reported as mean $\pm$ standard deviation over 10 evaluation runs with fixed checkpoints. SR is task success rate (\%), and SA is preference satisfaction (\%); No-Pref reports SR only. All variants start from the $\pi_{0.5}$ base checkpoint. All-Pref evaluates one checkpoint jointly trained across Motion, Skill, and User-Profile tasks. Mix-LoRA and Seq-LoRA additionally use LIBERO data: Seq-LoRA first fine-tunes on LIBERO and then on \suiteSinglePref{}, while Mix-LoRA fine-tunes on their mixture. $^*$ indicates the same task and preference under an unseen layout.}
    \label{tab:single_performance_sr}
    \label{tab:single_performance_sat}

    \makebox[\textwidth][c]{%
    \begin{tabular}{l|c|cc|cc|cc|cc|cc}
        \hline
        \multirow{2}{*}{\textbf{Method}}
            & \textbf{No-Pref}
            & \multicolumn{2}{c|}{\textbf{All-Pref}}
            & \multicolumn{2}{c|}{\textbf{\prefMotion{}}}
            & \multicolumn{2}{c|}{\textbf{\prefSkill{}}}
            & \multicolumn{2}{c|}{\textbf{\prefUserProfile{}}}
            & \multicolumn{2}{c}{\textbf{\suiteComprePref{}}} \\ \cline{2-12}
            & SR & SR & SA & SR & SA & SR & SA & SR & SA & SR & SA \\ \hline
        $\pi_{0.5}$-LoRA
            & - & \shortstack{\rule{0pt}{9pt}84.96\\\textpm{}1.06} & \shortstack{82.48\\\textpm{}1.02} & \shortstack{90.00\\\textpm{}2.85} & \shortstack{85.37\\\textpm{}2.66} & \shortstack{78.86\\\textpm{}1.44} & \shortstack{76.67\\\textpm{}1.40} & \shortstack{86.67\\\textpm{}1.49} & \shortstack{85.22\\\textpm{}1.47} & \shortstack{44.00\\\textpm{}0.71} & \shortstack{42.64\\\textpm{}0.69} \\[3pt]
        $\pi_{0.5}$-Mix-LoRA
            & \shortstack{91.00\\\textpm{}0.70} & \shortstack{84.55\\\textpm{}1.94} & \shortstack{84.45\\\textpm{}2.04} & \shortstack{66.88\\\textpm{}3.56} & \shortstack{66.06\\\textpm{}4.38} & \shortstack{89.29\\\textpm{}2.82} & \shortstack{89.29\\\textpm{}2.82} & \shortstack{91.33\\\textpm{}3.42} & \shortstack{91.33\\\textpm{}3.42} & \shortstack{72.83\\\textpm{}0.95} & \shortstack{70.07\\\textpm{}0.91} \\[3pt]
        $\pi_{0.5}$-Seq-LoRA
            & \shortstack{52.33\\\textpm{}0.37} & \shortstack{74.75\\\textpm{}1.97} & \shortstack{74.73\\\textpm{}2.12} & \shortstack{67.50\\\textpm{}5.68} & \shortstack{67.44\\\textpm{}6.52} & \shortstack{72.14\\\textpm{}2.04} & \shortstack{72.14\\\textpm{}2.04} & \shortstack{81.00\\\textpm{}2.79} & \shortstack{81.00\\\textpm{}2.79} & \shortstack{70.00\\\textpm{}1.32} & \shortstack{68.25\\\textpm{}1.29} \\[3pt]
        $\pi_{0.5}$-LoRA$^*$
            & - & \shortstack{37.49\\\textpm{}3.27} & \shortstack{28.36\\\textpm{}2.55} & \shortstack{55.00\\\textpm{}7.53} & \shortstack{43.43\\\textpm{}7.02} & \shortstack{28.57\\\textpm{}3.79} & \shortstack{26.94\\\textpm{}3.57} & \shortstack{34.00\\\textpm{}5.60} & \shortstack{20.58\\\textpm{}3.39} & - & - \\[3pt]
        \hline
    \end{tabular}
    }
\end{table*}

\noindent\textbf{\textit{Finding.}} Sequential training degrades no-preference and preference-conditioned performance. Mixed training instead yields a category-dependent trade-off: it better retains no-preference behavior and contextual-prompt robustness, but weakens motion preference relative to preference-only LoRA.

Table~\ref{tab:single_performance_sr} compares the No-Pref and All-Pref settings.
$\pi_{0.5}$-Seq-LoRA obtains $52.33{\pm}0.37\%$ No-Pref SR and $74.75{\pm}1.97\%$ All-Pref SR, whereas $\pi_{0.5}$-Mix-LoRA retains $91.00{\pm}0.70\%$ No-Pref SR and reaches $84.55{\pm}1.94\%$ All-Pref SR.
The \suiteComprePref{} columns further evaluate robustness to contextual prompt variation: Mix-LoRA reaches $72.83{\pm}0.95\%$ SR and $70.07{\pm}0.91\%$ SA, compared with $44.00{\pm}0.71\%$ and $42.64{\pm}0.69\%$ for preference-only LoRA. However, Mix-LoRA reduces Motion Preference SR from $90.00{\pm}2.85\%$ to $66.88{\pm}3.56\%$. The results therefore compare strategies for jointly learning preference-conditioned and preference-agnostic behavior; they do not establish mixed training as uniformly superior.

\textbf{Q3: Learning Efficiency.}
\label{sec:rq3}

\begin{figure}[!t]
    \centering
    \begin{subfigure}[t]{0.49\linewidth}
        \centering
        \includegraphics[width=\linewidth]{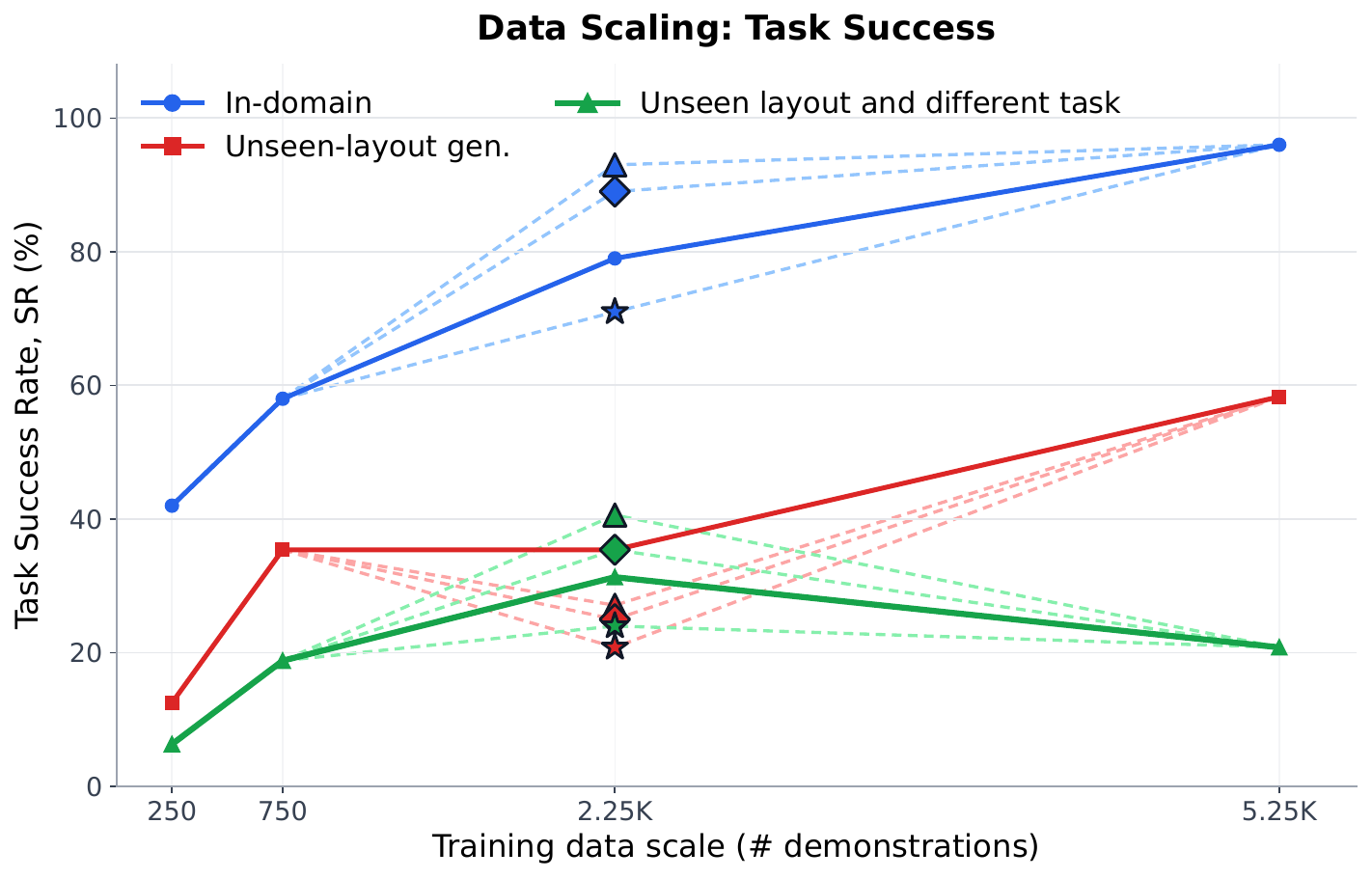}
    \end{subfigure}
    \hfill
    \begin{subfigure}[t]{0.49\linewidth}
        \centering
        \includegraphics[width=\linewidth]{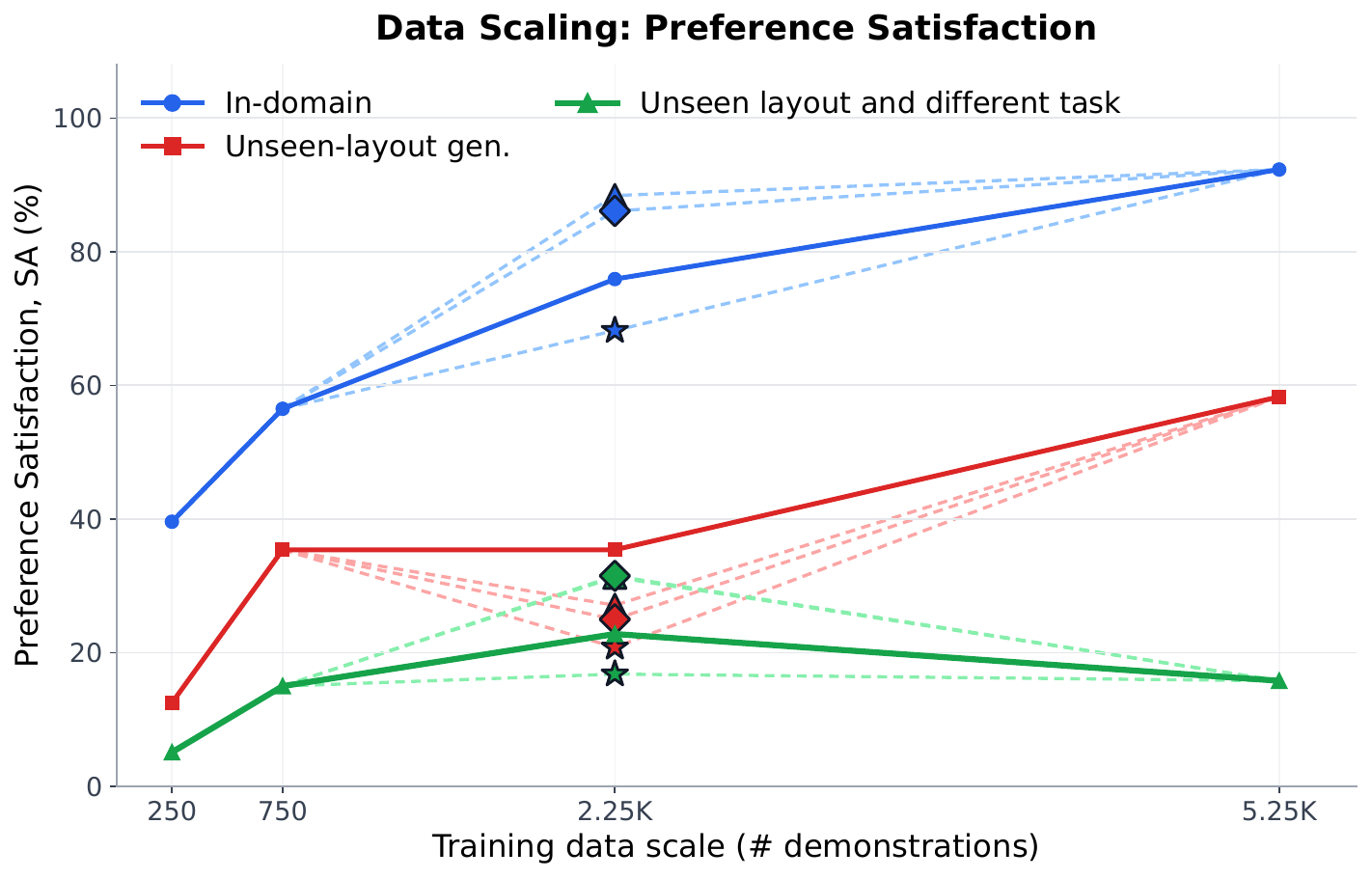}
    \end{subfigure}
    \captionsetup{font=normalsize}
    \caption{Preference scaling behavior under increasing data scale.
We report task success rate (SR) for in-domain (blue), unseen-layout (red), and unseen-layout-with-different-task (green) generalization.
Solid lines show the main scaling trend, while dashed branches compare L-series alternatives at the same 2.25K scale:
L1 ($\star$) emphasizes demonstration density over fewer layouts, L2 is the solid-line intermediate setting, L3 ($\triangle$) emphasizes broader layout coverage with fewer demonstrations per layout, and L4 ($\diamond$) adds matched no-preference missions for preference-layout decoupling. As the data scale increases, the number of training layouts also increases to provide broader layout coverage.}
    \label{fig:learning_efficiency}
\end{figure}

This question evaluates how data scale affects learning efficiency under the same task and preference distribution.
In Figure~\ref{fig:learning_efficiency}, the blue curves show in-domain scaling as the number of training demonstrations increases.
At the 2.25K scale, L1-L4 isolate data-composition effects under a comparable budget.
L1, L2, and L3 use $K=2$, $K=3$, and $K=5$ layouts, respectively, trading off layout diversity against demonstrations per layout.
L4 uses the same $K=3$ layout setting as L2 but adds matched no-preference missions, testing whether no-preference data helps decouple preference-conditioned behavior from task/layout cues.

\noindent\textbf{\textit{Finding.}}
Across all data scales, in-domain preference-conditioned behavior is learned consistently well.
For in-domain preference learning, increasing layout diversity is more effective than simply increasing the number of demonstrations.
Notably, adding matched no-preference data further improves learning even without increasing layout diversity, suggesting that such data helps decouple preference-conditioned behavior from task/layout-specific cues.

Figure~\ref{fig:learning_efficiency} provides detailed evidence for these observations.
First, the in-domain SR and SA curves are closely aligned across data scales, indicating strong in-domain preference learning: successful executions usually also satisfy the specified preferences.
Second, at the fixed 2.25K scale, L1-L3 use the same demonstration budget but increase the number of layouts, and the improvement from L1 to L3 shows that broader layout coverage is more effective than denser demonstrations within fewer layouts.
Third, L4 uses the same layout setting as L2 but adds matched no-preference missions; its improvement over L2 suggests that no-preference data benefits learning beyond increasing layout diversity, likely by helping separate task execution from preference-conditioned behavior.

\textbf{Q4: Generalization.}
\label{sec:rq4}
This question evaluates two forms of generalization:  task-and-layout generalization, where either the task or layout changes, and preference-level generalization, where both tasks and layouts change while the preference remains unchanged.
In Figure~\ref{fig:learning_efficiency}, the red curves denote unseen-layout generalization, and the green curves denote unseen-layout-with-different-task generalization. Both settings use the same L1-L4 controlled comparisons at the matched 2.25K scale.

\noindent\textbf{\textit{Finding.}}
Preference-conditioned behavior is consistently learned across settings, but jointly achieving task success and preference satisfaction remains challenging under generalization.
Unseen-layout generalization benefits from increased data scale, whereas generalization to both unseen layouts and different tasks remains limited. As shown in Figure~\ref{fig:learning_efficiency}, SR and SA are closely aligned within each setting, suggesting that successful executions usually also satisfy the specified preference.
However, under generalization, especially unseen-layout-with-different-task, both metrics remain much lower than in-domain performance.
This indicates that learning preference-conditioned behavior is easier than executing it successfully together with the task under new task-layout combinations.

\textbf{Q5: Composition and Conflict.}
\label{sec:rq5}
This question evaluates whether policies can compose preferences, decompose a target preference from joint training, and handle preference-preference or task-preference conflicts.
Tables~\ref{tab:preference_composition} and~\ref{tab:preference_conflict} report the results.
For composition/decomposition, Eval SR/SA denote task success and target preference satisfaction on the evaluation condition.
For conflicts, Pref-Dom.\ measures the gap between following either preference under a joint prompt; lower values indicate less consistent preference domination.
In task-preference conflicts, Follow Task, Follow Preference, and Incomplete denote completing the task while violating the preference, following the preference while violating the task, and satisfying neither, respectively.

\noindent\textbf{\textit{Finding.}}
Preference composition is partially learnable, whereas decomposition fails. After learning $P_A$ and $P_B$ separately, the policy can compose them into $P_A{+}P_B$ with 25.0\% Eval SR/SA.
However, when asked to infer $P_B$ from $P_A{+}P_B$ and $P_A$, the policy obtains 37.5\% Eval SR but 0.0\% Eval SA, showing task success does not imply successful preference behavior execution.

\noindent\textbf{\textit{Finding.}}
Conflicting preferences reduce performance and have no dominant preference. Under preference-preference conflicts, conflict-time SR/SA drop to 66.00\%/47.75\%, and Pref-Dom. is only 11.25\%, suggesting no clear preference dominates the other under our either-preference success rule.
Under task-preference conflicts, Follow Task and Follow Preference are close, 33.3\% vs. 29.2\%, with 37.5\% incomplete trials, indicating that conflicts often lead to failed execution rather than task- or preference-following behavior.

\FloatBarrier
\begin{table*}[!t]
    \centering
    \footnotesize
    \begin{minipage}[t]{0.48\textwidth}
        \vspace{0pt}
        \centering
        \renewcommand{\arraystretch}{1.15}
        \setlength{\tabcolsep}{2.4pt}
        \caption{Preference composition tests forming $P_A + P_B$ from $P_A$ and $P_B$; decomposition tests inferring $P_B$ from $P_A + P_B$ and $P_A$.}
        \label{tab:preference_composition}

        \makebox[\linewidth][c]{%
        \begin{tabular}{@{}lcccc@{}}
            \toprule
            \multicolumn{5}{@{}l}{\textit{Preference Composition}} \\
            \textbf{Eval} & \shortstack{\textbf{In-dom.}\\\textbf{SR}} & \shortstack{\textbf{In-dom.}\\\textbf{SA}} & \shortstack{\textbf{Eval}\\\textbf{SR}} & \shortstack{\textbf{Eval}\\\textbf{SA}} \\
            \midrule
            $P_A{+}P_B$ & 58.3 & 58.3 & 25.0 & 25.0 \\
            \midrule
            \multicolumn{5}{@{}l}{\textit{Preference Decomposition}} \\
            \textbf{Eval} & \shortstack{\textbf{In-dom.}\\\textbf{SR}} & \shortstack{\textbf{In-dom.}\\\textbf{SA}} & \shortstack{\textbf{Eval}\\\textbf{SR}} & \shortstack{\textbf{Eval}\\\textbf{SA}} \\
            \midrule
            $P_B$ & 66.7 & 66.7 & 37.5 & 0.0 \\
            \bottomrule
        \end{tabular}
        }
    \end{minipage}
    \hfill
    \begin{minipage}[t]{0.48\textwidth}
        \vspace{0pt}
        \centering
        \renewcommand{\arraystretch}{1.15}
        \setlength{\tabcolsep}{2.4pt}
        \caption{Conflict performance. Preference-preference success requires completing the task while satisfying one of the two conflicting preferences; task-preference conflicts prioritize the task.}
        \label{tab:preference_conflict}

        \makebox[\linewidth][r]{%
        \begin{tabular}{@{}lccccc@{}}
            \toprule
            \multicolumn{6}{@{}l}{\textit{Preference-Preference Conflict}} \\
            \textbf{Eval} & \shortstack{\textbf{In-dom.}\\\textbf{SR}} & \shortstack{\textbf{In-dom.}\\\textbf{SA}} & \shortstack{\textbf{Conf.}\\\textbf{SR}} & \shortstack{\textbf{Conf.}\\\textbf{SA}} & \shortstack{\textbf{Pref}\\\textbf{Dom.}} \\
            \midrule
            $P_1 + P_2$ & 73.00 & 63.40 & 66.00 & 47.75 & 11.25 \\
            \midrule
            \multicolumn{6}{@{}l}{\textit{Task-Preference Conflict}} \\
            \textbf{Eval} & \shortstack{\textbf{In-dom.}\\\textbf{SR}} & \shortstack{\textbf{In-dom.}\\\textbf{SA}} & \shortstack{\textbf{Follow}\\\textbf{task (SR)}} & \shortstack{\textbf{Follow}\\\textbf{pref.}} & \textbf{Incomplete} \\
            \midrule
            $T_A + P_B$ & 50.00 & 50.00 & 33.3 & 29.2 & 37.5 \\
            \bottomrule
        \end{tabular}
        }
    \end{minipage}
\end{table*}

\vspace{-0.2cm}
\section{Conclusion}
\vspace{-0.2cm}
\label{sec:conclusion}
In this work, we introduce \toolname{}, a comprehensive benchmark for personalized robotic manipulation. By treating preferences as a first-class building block alongside tasks and layouts, \toolname{} provides five task suites that isolate and evaluate preference learning under single-preference, scaling, conflict, composition, and comprehensive preference reasoning settings.
Experiments show that current VLA policies struggle with zero-shot personalization, while SFT makes preferences learnable. Compared with sequential adaptation, mixed training better preserves no-preference behavior and improves contextual-prompt robustness, but it weakens motion-preference performance relative to preference-only LoRA.
Preference learning benefits more from layout diversity and same-layout no-preference data than denser demonstrations, but generalization to unseen layouts and different tasks remains difficult.
Composition is only partially observed, decomposition fails, and conflicts often degrade execution instead of producing resolution.

\textbf{Limitations and Future Work.} \toolname{} instantiates predefined, language-expressible, automatically measurable preferences in simulation. It therefore excludes implicit preference inference, interaction history, non-linguistic signals, and open-ended or evolving preferences. Its generated demonstrations may contain pipeline- and skill-library-specific biases despite randomized initial states, multiple missions per preference, and human review. The current composition suite combines only two preferences, and the conflict suite does not test conflict detection or clarification. Finally, the human survey provides only initial validation; no real-robot or longitudinal user study is included. Future work could extend \toolname{} in directions ranging from three-or-more-preference settings to non-linguistic preference signals, human-in-the-loop preference acquisition, preference weighting and when-to-ask decisions, and real-robot evaluation.

\bibliographystyle{unsrtnat}
\bibliography{references}

\clearpage
\appendix
\section{Additional Information}
\label{app:additional_experiments}

This appendix includes additional experimental details and results that support the main findings in Section~\ref{sec:experimental_results}.

\subsection{Ethics Statement}
\label{app:ethics}

\toolname{} is built from procedurally generated preference-conditioned tasks, layouts, trajectories, and evaluation signals in simulation. Human feedback is used to review generated data and validate preference plausibility, but no personal data or real-world user logs are released or used as policy-training data. Its intended use is research on personalized robotic manipulation, including preference learning, generalization, composition, and conflict handling.
Our dataset construction process is designed to reduce ethical and legal risk. We build on LIBERO~\cite{liberoliu2023} scenes and simulator assets, then generate preference-conditioned variants through minimal layout edits, procedural trajectory generation, and automatic preference metrics. Preference descriptions, satisfaction rules, and benchmark annotations are created for research use rather than copied from external products or commercial systems. Any public release should preserve attribution to upstream assets and follow the licenses and terms of use of LIBERO and other dependencies.
The benchmark also has important limitations. Our preference metrics only capture observable execution outcomes in simulation and cannot represent open-ended, private, or highly subjective human preferences. As a result, \toolname{} should not be treated as a substitute for human evaluation in safety-critical or real-world personalized deployment. We intend this benchmark to support controlled scientific evaluation, not direct real-world decision making.

We take responsibility for the released benchmark artifacts. The \toolname{} dataset and open-source code are released under the MIT license.

\subsection{Training Details}
\label{app:training_details}

For non-zero-shot experiments, we further tune $\pi_{0.5}$ with LoRA~\cite{hu2022lora}. We scale the training budget with the number of rollouts: for every 4,000 rollout episodes, we use roughly 45,000 optimization steps on eight A6000 Ada GPUs with a learning rate of $5\times10^{-5}$.
For the main fine-tuning comparison, All-Pref denotes one checkpoint jointly trained on Motion, Skill, and User-Profile preference tasks; it is not an ensemble or an average of category-specific checkpoints. We evaluate each fixed checkpoint in 10 independent runs and report mean $\pm$ standard deviation in Table~\ref{tab:single_performance_sr}. The variation therefore reflects randomized evaluation rollouts, not training seeds.

\section{Preference Selection Criteria}
\label{app:preference_selection}

We select a preference only if it satisfies five criteria: 1) expressible in natural language; 2) can be grounded in existing LIBERO~\cite{liberoliu2023} scenes; 3) satisfiable without changing the task goal; 4) measurable by an automatic preference metric; and 5) reusable across multiple tasks or layouts.
Examples include the \prefMotion{} preference \textit{``speed up''}, measured by execution steps; the \prefSkill{} preference \textit{``pick up an object by its handle''}, measured by waypoints and contact points; and the \prefUserProfile{} preference \textit{``prefer dairy product''}, measured by the executed subgoal.
We exclude preferences that require scene-external information, unsafe or task-infeasible behavior, or subjective judgments that cannot be mapped to observable execution outcomes.

Table~\ref{tab:single_preference_catalog} lists all 37 single-preference instances in \suiteSinglePref{}.

\begin{table}[p]
    \centering
    \small
    \caption{Complete list of the 37 single-preference instances in \suiteSinglePref{}.}
    \label{tab:single_preference_catalog}
    \renewcommand{\arraystretch}{1.05}
    \begin{tabular}{@{}lp{0.82\linewidth}@{}}
        \toprule
        \textbf{ID} & \textbf{Preference} \\
        \midrule
        \multicolumn{2}{l}{\textbf{Motion (8 preferences)}} \\
        1-1 & Speed up \\
        1-2 & Slow down \\
        1-3 & Avoid sudden acceleration \\
        1-4 & Keep away from wine bottle / Keep away from stove / Keep away from the other bowl \\
        1-5 & Close to ground when moving \\
        1-6 & Move straight \\
        1-7 & Move higher \\
        1-8 & Rotate when moving \\
        \midrule
        \multicolumn{2}{l}{\textbf{Skill (14 preferences)}} \\
        2-1 & Pick from right edge \\
        2-2 & Pick from back edge \\
        2-3 & Pick by handle \\
        2-4 & Pick from top at an angle \\
        2-5 & Pick from left side \\
        2-6 & Pick from right side \\
        2-7 & Pick from horizontal side \\
        2-8 & Push instead of pick-and-put \\
        2-9 & Drag instead of pick-and-put \\
        2-10 & Throw instead of placing \\
        2-11 & Pick from front \\
        2-12 & Rotate when placing \\
        2-13 & Pick from left edge \\
        2-14 & Lay object down when placing \\
        \midrule
        \multicolumn{2}{l}{\textbf{User Profile (15 preferences)}} \\
        3-1 & Prefer green objects \\
        3-2 & Prefer red objects \\
        3-3 & Prefer yellow objects \\
        3-4 & Prefer blue objects \\
        3-5 & Prefer bigger objects \\
        3-6 & Upper drawer for food and drink \\
        3-7 & Middle drawer for kitchenware \\
        3-8 & Serve double portion \\
        3-9 & Prefer dairy products \\
        3-10 & Prefer sweet drinks \\
        3-11 & Prefer sour sauces \\
        3-12 & Prefer boxed objects \\
        3-13 & Prefer bottled objects \\
        3-14 & Prefer canned food \\
        3-15 & Prefer smaller objects \\
        \midrule
        \multicolumn{2}{l}{\textbf{Total (37 preferences)}} \\
        \bottomrule
    \end{tabular}
\end{table}

\FloatBarrier

\section{Model-Assisted Generation and Validation}
\label{app:generation_details}

We use Claude Code with an Opus-family model and Gemini as backends for three model-assisted stages. In \emph{layout editing}, the model receives a BDDL scene specification, preference language, object geometry, and spatial constraints, and proposes a modified BDDL layout. In \emph{personalized planning}, it receives the original task and preference instructions, baseline plan, available objects and skills, and a scene image, and returns a sequence of (skill, object) pairs plus required, prohibited, and ordering constraints. In \emph{image inspection}, independent VLM judgments assess task completion and visually observable preference compliance from rollout frames.

Inspired by single- and multi-agent evaluations~\cite{sun2026agents,wang2026coopuq,11696967}, the layout and plan candidates follow a generate-and-trial loop: each proposal is instantiated and tested in simulation, and any candidate violating the output format, physical constraints, task-success condition, or preference constraints is rejected. The observed failure is included when generating a revised candidate. The reported VLM failure rate in each stage was below 23\%. For image inspection, inconsistent independent judgments are flagged for re-evaluation rather than accepted directly. Image judgments are restricted to visible properties such as contact location, approach direction, gripper orientation, and final placement. Motion speed, acceleration, trajectory smoothness, and clearance height are instead measured with trajectory-based geometric or kinematic metrics.

The following condensed prompts specify the inputs, output formats, and constraints used in the three stages. Complete prompts, model configurations, and validation logic will accompany the code.

{\small
\begin{verbatim}
[Layout Editing]
Generate an initial position for {obj_name} with size
{obj_dimensions} under constraint "{lan_description}".
Existing objects: {name}: pos=({x},{y}), size=({w},{h}), ...
Workspace: x,y in [-0.25,0.25].
Choose collision-free (x,y) satisfying the constraint,
reachability, workspace margins, >=0.03m clearance,
and layout diversity.
Guidelines: accessible/center/reachable: within 0.10m of (0,0);
clustered with X: 0.02-0.04m beyond collision;
left/right: x<-0.05/x>0.05; front/back: y>0.05/y<-0.05.
Return only: {"x":float,"y":float,"reasoning":str}

[Personalized Planning]
Task: {lan_inst}
Baseline: {task_plan.plans}
Preference: {high_level_pref}
Existing objects: {layout.object_set}
Available objects: {available_obj_set}
Skills: {high_skill_name_and_desc}
Preserve the task goal and ground the preference into
(skill,object) pairs. Add 1-5 objects only if needed and
use geometry-compatible skills.
Return: (1) personalized sequence;
(2) "required", "prohibited", "order" constraints;
(3) unchanged preference.
Wildcards: ("skill","_ALL")=any object;
("_ALL","object")=any skill; ("skill","object")=specific pair.
Format:
1.||('skill','object'),...||
2.||"required":[...],"prohibited":[...],"order":[...]||
3.||{high_level_pref}||
Generate {n_task_per_response} plans.
Retry feedback: {additional_prompt}

[Image Inspection]
Evaluate robot-manipulation frames in reading order and
independently judge task success and preference compliance.
Goal: "{main_goal}"
Preference: "{pref_lan}"
Baseline: {naive_plan_text}
Preference plan: {task_plan_text}
Subtasks/frames: {steps_text}
Skill semantics: {base_skill_description}
Preference semantics: {pref_variant_description}
Use only visible evidence: contact location, gripper orientation,
approach direction, and final pose. Do not infer speed,
acceleration, smoothness, or clearance height; return
"unverifiable" when compliance is not visually observable.
Judge completion for each subtask and compliance for each
preference-relevant subtask.
Output:
Task_Verdict:<yes|no>
Task_Reason:<brief justification>
Preference_Verdict:<yes|no|unverifiable>
Preference_Reason:<brief justification with frame numbers>
Subtask_Analysis:
Subtask 1: completed=<yes|no>|
preference=<yes|no|n/a|unverifiable>-<reason>
...
\end{verbatim}
}

\section{Human Review and Initial User Validation}
\label{app:human_validation}

\noindent\textbf{Construction and cross-review.}\par
Human contributors participated in three stages: 1) providing one reference demonstration for each newly introduced skill; 2) inspecting sampled generated demonstrations together with their satisfaction labels and metrics; and 3) rating preference plausibility and demonstration human-likeness. Before beginning, contributors received common instructions and task-specific training. Contributors were assigned to different portions of the generated data, after which a second group cross-reviewed the demonstrations and labels. Disputed cases were jointly re-examined before inclusion; the initial reviewer disagreement rate was 11.8\%. All participants provided informed consent, and contributors were credited according to their level of involvement.

\noindent\textbf{Survey protocol and results.}\par
For initial user validation, 15 participants rated the selected preferences and their demonstrations on a 10-point scale along two dimensions: \emph{real-world preference plausibility}, whether each preference was plausible and applicable in a household setting; and \emph{human-likeness of demonstrations}, how closely the corresponding demonstration resembled how a human would naturally perform the task while satisfying the preference. Table~\ref{tab:user_validation} summarizes 37 preferences in three categories. These results provide initial evidence of plausibility and human-likeness, but they do not substitute for real-robot, longitudinal, or larger-scale user studies.

Before participating, survey respondents were informed of the study's purpose and procedure and provided informed consent. They received compensation via WeChat in accordance with fair labor practices. We did not identify any risks beyond those ordinarily associated with completing a questionnaire.

\begin{table}[htbp]
    \centering
    \small
    \caption{Initial human validation with 15 participants; ratings use a 10-point scale.}
    \label{tab:user_validation}
    \begin{tabular}{lccc}
        \toprule
        \textbf{Preference type} & \textbf{Instances} & \textbf{Human-likeness} $\uparrow$ & \textbf{Plausibility} $\uparrow$ \\
        \midrule
        Motion & 8 & 8.63 & 9.25 \\
        Skill & 14 & 8.50 & 8.50 \\
        User Profile & 15 & 9.93 & 7.73 \\
        \midrule
        Overall & 37 & 9.11 & 8.35 \\
        \bottomrule
    \end{tabular}
\end{table}

\FloatBarrier
\section{A Worked Example of ComprePref}
\label{app:comprepref_example}

\begin{center}
{\setlength{\fboxsep}{7pt}
\fbox{\begin{minipage}{0.94\linewidth}
\small
\textbf{Underlying task:} Put the white mug on the left plate.\par
\textbf{Explicit preference:} Pick by the handle.\par
\textbf{Input prompt:} Put the white mug on the left plate. The mug just came out of the microwave---the body is too hot to touch.\par
\textbf{Implied behavior:} Grasp the handle while completing the same placement task.\par
\textbf{What stays fixed:} The task goal, scene, and preference evaluator. SR evaluates the original placement goal; SA uses the original handle-grasp preference label with task-success gating.\par
\textbf{What changes:} The language supplied to the policy conveys the preference through context instead of a separate explicit preference field.
\end{minipage}}}
\end{center}

A second example replaces the explicit preference ``speed up'' with ``Guests will use them immediately'' for the task of putting alphabet soup and a cream cheese box in a basket. The motion preference is still evaluated from the trajectory, not from a still image. These contextual prompts assess grounding of a specified benchmark preference; they do not establish recovery of an individual user's unobserved preference history.

\section{Adding a New Preference}
\label{app:preference_extensibility}

Adding a standard preference requires one human reference demonstration and a natural-language description of its satisfaction metric for the preference-conditioned generation pipeline, followed by approximately three to five validation trials for each applicable task. For a preference applied to roughly five tasks, the observed manual effort is about 20 minutes on average; demonstration generation then proceeds automatically. This estimate covers preferences expressible through the current scene representation, skill library, and automatic programmable metrics. Preferences requiring a new skill or non-observable feedback incur additional work.

For example, to add a preference for grasping a mug by its handle, a user provides one reference demonstration and a natural-language description of the satisfaction metric: the robot should grasp the handle rather than the mug body. The original mug-placement goal remains unchanged. The user then runs approximately three to five validation trials for each applicable task to check whether the generated trajectories follow the preference and whether the metric correctly evaluates grasp location. After validation, the pipeline automatically generates additional demonstrations.

\section{Benchmark Release and Maintenance}
\label{app:release}

The benchmark has five evaluation suites. The 312 tasks include both preference-conditioned and no-preference tasks across SinglePref (100), PrefScale (100), PrefConflict (48), and PrefCompose (64); ComprePref reuses tasks with contextual prompts and does not add new mission-preference combinations to this count. Demonstrations are generated trajectories used as training/reference data; evaluation rollouts are executions of the evaluated policy and are not additional benchmark tasks.

The release documentation covers installation, dataset/configuration descriptions, and examples for generation, training, and inference. The configuration follows the five-suite organization described here, and the Croissant metadata includes the fields \texttt{rai:personalSensitiveInformation} and \texttt{rai:hasSyntheticData}. We will maintain the code, datasets, and leaderboard, improve the documentation based on community feedback, and expand rollout coverage and generation diversity.

\end{document}